\documentclass{esannV2}
\usepackage[dvips]{graphicx}
\usepackage{xcolor}
\usepackage[latin1]{inputenc}
 \usepackage{booktabs}
\usepackage{amsthm}

\usepackage{amssymb,amsmath,array}
\usepackage{tikz}
\usetikzlibrary{shapes.geometric,positioning,calc,decorations.pathreplacing,arrows.meta}
\usetikzlibrary{positioning,calc,decorations.pathreplacing,arrows.meta}
\usetikzlibrary{arrows.meta,decorations.pathmorphing,backgrounds,positioning,fit,petri}
\usepackage{lineno}

\theoremstyle{definition}
\newtheorem{theorem}{Theorem}%[section]
\newtheorem{corollary}{Corollary}[theorem]
\newtheorem{lemma}[theorem]{Lemma}
\newtheorem{definition}{Definition}[section]
\usepackage{amsmath,amsfonts,bm}

\def\eqref#1{equation~\ref{#1}}
\def\1{\bm{1}}

\def\vone{{\bm{1}}}

\def\vb{{\bm{b}}}

\def\mC{{\bm{C}}}
\def\mD{{\bm{D}}}

\def\mH{{\bm{H}}}
\def\mI{{\bm{I}}}

\def\mP{{\bm{P}}}

\def\mS{{\bm{S}}}
\def\mT{{\bm{T}}}
\def\mU{{\bm{U}}}
\def\mV{{\bm{V}}}
\def\mW{{\bm{W}}}
\def\mX{{\bm{X}}}

\def\mZ{{\bm{Z}}}

\DeclareMathAlphabet{\mathsfit}{\encodingdefault}{\sfdefault}{m}{sl}
\SetMathAlphabet{\mathsfit}{bold}{\encodingdefault}{\sfdefault}{bx}{n}

\renewcommand{\eqref}[1]{(\ref{#1})}

\newcommand{\op}[1]{|| #1 ||_{op}}
\newcommand{\fr}[1]{|| #1 ||_F}

\newcommand{\extref}{the appendix}
\newcommand{\appref}[2]{\ref{#1}}

\begin{document}
% \linenumbers
%style file for ESANN manuscripts
\title{On the Informativeness of Structural Stability Bounds for GNNs:\\GINs versus GATs\\(Extended Version)}

%***********************************************************************
% AUTHORS INFORMATION AREA
%***********************************************************************
\author{Alan Gany$^1$, Paul Landrier$^2$, Bogdan Cautis$^{3,1}$, Laks V. S. Lakshmanan$^4$
and Silviu Maniu$^5$
\thanks{Extended version of a paper submitted to ESANN 2027.}
%
% DO NOT MODIFY THE FOLLOWING '\vspace' ARGUMENT
\vspace{.3cm}\\
1- Universit\'e Paris-Saclay, CNRS, LISN - France
\vspace{.1cm}\\
2- \'Ecole normale sup\'erieure - France
\vspace{.1cm}\\
3- Singapore Institute of Technology - Singapore
\vspace{.1cm}\\
4- University of British Columbia - Vancouver, Canada
\vspace{.1cm}\\
5- Univ. Grenoble Alpes, CNRS, Grenoble INP, LIG - France\\
}
%***********************************************************************
% END OF AUTHORS INFORMATION AREA
%***********************************************************************

\maketitle

\begin{abstract}
Stability bounds for graph neural networks limit how much node
representations can drift when the graph changes, but they are rarely
compared with the simplest alternative: a bound on the distance between any
two outputs, whatever the graph. We derive structural stability bounds for
Graph Isomorphism Networks (GINs) and single-head GATv1 networks under
deterministic edge perturbations that need not be small, and compare each
with such a perturbation-independent diameter bound. For GINs, we
characterize exactly when the stability bound is tighter, through a
condition linking perturbation size, depth and maximum degree; it holds with
a wide margin on Cora, but fails beyond a few layers on the low-degree MUTAG
graphs. For GATv1, the bound is never tighter, which calls for analyses
based on attention masses.
\end{abstract}

\section{Introduction and Motivation}

Graph neural networks (GNNs) are increasingly deployed on graphs that are
noisy, evolve over time, or may be manipulated, so it matters how much their
node representations can change when edges change. Stability theory answers
this with upper bounds on the representation drift caused by a perturbation
of the graph. Such bounds have mostly been derived for convolutional and
spectral architectures under deterministic perturbations
\cite{gama2020stability,kenlay2021stabilitygcn,koke2024transferability},
with extensions to broader families under small relative perturbations
\cite{gao2023tradeoff}.

Stability bounds are worst-case guarantees, and a guarantee is only useful if
it says more than what holds trivially. For fixed input features and model
parameters, the outputs of a GNN over all admissible graphs lie in a bounded
set, so the drift caused by \emph{any} perturbation is at most the diameter
of that set. A stability bound that exceeds a bound on this diameter is
therefore vacuous: it uses the perturbation, yet tells less than a bound that
ignores it. We call a stability bound \emph{informative} when it is strictly
tighter than such a perturbation-independent diameter bound, and ask when
this is the case.

We study two architectures that have received less attention than
convolutional ones. For Graph Isomorphism Networks (GINs)
\cite{xu2018powerful}, we derive a bound that explicitly accounts for the
Lipschitz constants of the MLPs and for the propagation of perturbations
across layers. For single-head GATv1 \cite{velivckovic2017graph}, we use the
$1/2$-Lipschitz property of softmax \cite{nair2025softmax} to control both
the change of aggregation neighborhoods and the change of the
representation-dependent attention coefficients. Our perturbations are
deterministic edge additions or deletions, not required to be small, with
features and parameters fixed. We then characterize exactly when each bound
is informative. For GINs, this holds under an explicit condition linking the
perturbation size, the depth and the maximum degree, which we evaluate on
Cora and MUTAG under random, structural and adversarial edge deletions. For
GATv1, the bound we obtain is never informative for a nonzero perturbation,
which identifies where the analysis must be sharpened.

\section{Preliminaries on Graph Neural Networks}\label{sec:GNN}

\noindent\textbf{Graphs.} A graph $\mathcal{G}=(V,E)$ has $N=|V|$ nodes and
edges $E\subseteq V\times V$. We represent it by its binary adjacency matrix
$S\in\{0,1\}^{N\times N}$, which is symmetric with zero diagonal since we
consider undirected graphs without self-loops; self-information is added by
each architecture. Input node features form a matrix
$X\in\mathbb{R}^{N\times F^{(0)}}$, whose $i$th row we write $X_i$.

\begin{definition}[GIN]\label{def:GIN}
An $L$-layer Graph Isomorphism Network (GIN) $\Phi$ with parameters
$(\mathrm{MLP}_\ell,\epsilon_\ell)_{\ell=0}^{L-1}$ maps $(S,X)$ to
$\Phi(S,X)=X^{(L)}$, where $X^{(0)}=X$ and
\[
X^{(\ell+1)}=\mathrm{MLP}_\ell\bigl(((1+\epsilon_\ell)I_N+S)\,X^{(\ell)}\bigr),
\qquad \ell=0,\ldots,L-1,
\]
with $\epsilon_\ell\in\mathbb{R}$ and each $\mathrm{MLP}_\ell$ applied row-wise.
\end{definition}

\begin{definition}[GATv1]\label{def:gat}
An $L$-layer, single-head, bias-free Graph Attention Network (GATv1) $\Phi$
has parameters $W^{(\ell)}\in\mathbb{R}^{F^{(\ell)}\times F^{(\ell+1)}}$
and $a^{(\ell)}\in\mathbb{R}^{2F^{(\ell+1)}}$, $\ell=0,\ldots,L-1$. Let
$\mathcal{N}_S^+(i):=\{j:S_{ij}=1\}\cup\{i\}$. Then
$\Phi(S,X)=X^{(L)}$, where $X^{(0)}=X$ and
\[
X^{(\ell+1)}=\sigma\bigl(P^{(\ell)}(S,X^{(\ell)})\,X^{(\ell)}W^{(\ell)}\bigr),
\]
with attention matrix
$P^{(\ell)}_{ij}(S,H)=\exp(e^{(\ell)}_{ij}(H))/\sum_{k\in\mathcal{N}_S^+(i)}\exp(e^{(\ell)}_{ik}(H))$
for $j\in\mathcal{N}_S^+(i)$ and $0$ otherwise, and scores
\[
e^{(\ell)}_{ij}(H)=\operatorname{LeakyReLU}\bigl((a^{(\ell)})^\top[\,(H_iW^{(\ell)})^\top\Vert(H_jW^{(\ell)})^\top\,]\bigr),
\]
where $\Vert$ denotes concatenation.
\end{definition}

% \section{State of the art}
% \alan{I commented the previous state of the art section as it doesn't add much about what we are working on, maybe the section on adversarial attacks and certified robustness could be still relevant}
% \input{sota}

\section{Stability of GNNs -- The main theorems}\label{sec:stability}

In this section, we state the two structural stability bounds and
characterize when each is informative. All proofs are given in \extref.

\paragraph{Setting.}
We compare the outputs of a GNN $\Phi$ on two graphs with adjacency
matrices $S,\widehat S\in\{0,1\}^{N\times N}$ on the same labelled vertex
set, both symmetric with zero diagonal. The input features $X$ and all
trainable parameters are identical in the two evaluations, and inference is
deterministic: there is no dropout and no normalization layer. The
perturbation is thus purely topological, but it is not required to be small.
All activations are $1$-Lipschitz and vanish at zero. The quantity we bound
is the \emph{geometric drift}
$\Delta_\Phi(S,\widehat S;X):=\|\Phi(S,X)-\Phi(\widehat S,X)\|_F$.

\paragraph{Informativeness.}
For fixed $X$ and parameters, let $\mathcal{R}_\Phi(X)$ be the set of
outputs $\Phi(S,X)$ over all admissible adjacency matrices $S$. The drift
caused by any perturbation is at most the diameter of this set, so we call a
stability bound \emph{informative} when it is strictly smaller than a
perturbation-independent upper bound on
$\operatorname{diam}(\mathcal{R}_\Phi(X))$.

\subsection{Stability of GINs}

\begin{theorem}[Stability of GINs]\label{thm:gin}
Let $\Phi$ be an $L$-layer GIN (Definition~\ref{def:GIN}), $L\geq1$, whose
$\ell$th MLP consists of $K_\ell$ bias-free linear maps
$W_{\ell,1},\ldots,W_{\ell,K_\ell}$ interleaved with elementwise activations.
Let $d_{\max}$ be the maximum node degree over both graphs, and define
\[
\lambda_\ell:=\prod_{j=1}^{K_\ell}\|W_{\ell,j}\|_{op},
\qquad
q_\ell:=|1+\epsilon_\ell|+d_{\max}.
\]
Then
\[
\|\Phi(S,X)-\Phi(\widehat S,X)\|_F
\leq
\|S-\widehat S\|_{op}\;
\Bigl(\prod_{\ell=0}^{L-1}\lambda_\ell\Bigr)
\Bigl(\sum_{t=0}^{L-1}\prod_{r\neq t}q_r\Bigr)\,\|X\|_F .
\]
\end{theorem}

Over all graphs with maximum degree at most $d_{\max}$, the representation
diameter satisfies
$\operatorname{diam}(\mathcal{R}_{\Phi}(X))\le
U_{\mathrm{diam}}^{\mathrm{GIN}}:=2\bigl(\prod_{\ell}\lambda_\ell q_\ell\bigr)\|X\|_F$
(Lemma~\appref{lem:gin_diameter}{8} in \extref). Comparing the two bounds
gives an exact condition.

\begin{corollary}[Informativeness for GINs]
\label{cor:gin_informativeness}
Under the assumptions of Theorem~\ref{thm:gin}, assume $\|X\|_F>0$ and
$\lambda_\ell,q_\ell>0$ for all $\ell$, and let $\delta=\|S-\widehat S\|_{op}$.
The bound of Theorem~\ref{thm:gin} is strictly smaller than
$U_{\mathrm{diam}}^{\mathrm{GIN}}$ if and only if
$\delta\sum_{\ell=0}^{L-1}q_\ell^{-1}<2$. In particular, if $q_\ell=q$ for
every layer, the condition reduces to
\begin{equation}\label{eq:gin_coll2}
\delta<\frac{2q}{L}.
\end{equation}
\end{corollary}

The condition does not depend on the weights or the features: a higher
maximum degree favors informativeness, whereas greater depth and larger
perturbations work against it.

\subsection{Stability of GATv1}

\begin{theorem}[Stability of GATv1]\label{thm:gatv1}
Let $\Phi$ be an $L$-layer, single-head, bias-free GATv1
(Definition~\ref{def:gat}), $L\geq1$, with LeakyReLU slope in $[0,1]$. Let
$d_{\max}:=\max_{T\in\{S,\widehat S\}}\max_i|\mathcal{N}_T^+(i)|$ be the
largest attention neighborhood, $\|X\|_{2,\infty}:=\max_i\|X_i\|_2$, and, for
each layer,
\[
w_\ell:=\|W^{(\ell)}\|_{op},\qquad
B_\ell:=\|X\|_{2,\infty}\prod_{t<\ell}w_t,\qquad
\gamma_\ell:=\sqrt{2N}\,w_\ell B_\ell,
\]
\[
q_\ell:=w_\ell\Bigl(\sqrt{d_{\max}}+\sqrt{\tfrac{Nd_{\max}}{2}}\,B_\ell\|a^{(\ell)}\|_2w_\ell\Bigr).
\]
Then
\[
\|\Phi(S,X)-\Phi(\widehat S,X)\|_F
\leq
\|S-\widehat S\|_F\,
\sum_{t=0}^{L-1}\gamma_t\prod_{r=t+1}^{L-1}q_r .
\]
\end{theorem}

The second term of $q_\ell$ accounts for attention coefficients that change
because the representations change, and $\gamma_\ell$ for those that change
because the neighborhoods change. The representation diameter satisfies
$\operatorname{diam}(\mathcal{R}_{\Phi}(X))\le
U_{\mathrm{diam}}^{\mathrm{GAT}}:=2\sqrt{N}\,B_L$
(Lemma~\appref{lem:gat_diameter}{19} in \extref).

\begin{lemma}[Informativeness for GATv1]
\label{lem:gat_informativeness}
Under the assumptions of Theorem~\ref{thm:gatv1}, assume $w_\ell>0$ for
every layer and $\|X\|_{2,\infty}>0$, and let $\delta=\|S-\widehat S\|_F$.
The bound of Theorem~\ref{thm:gatv1} is strictly smaller than
$U_{\mathrm{diam}}^{\mathrm{GAT}}$ if and only if
$\delta\sum_{t=0}^{L-1}\prod_{r=t+1}^{L-1}q_r/w_r<\sqrt2$. Consequently, it
is not informative for any nonzero admissible perturbation.
\end{lemma}

The reason is structural: any edge flip gives $\delta\ge\sqrt2$, and the
last term of the sum equals one, so the topological contribution alone
already reaches the diameter bound. This contribution charges every row of
the attention matrix whose neighborhood changes with the largest possible
change, regardless of how much attention the added or removed neighbors
actually receive.

\section{Experiments}
\paragraph{Setup.}
We examine when the GIN bound is informative, as characterized by
Corollary~\ref{cor:gin_informativeness}. We take $\epsilon_\ell=0$, so that
Eq.~\eqref{eq:gin_coll2} reads $\delta L<2(1+d_{\max})$ with
$\delta=\|S-\widehat S\|_{op}$. The ratio
$r=U_{\mathrm{stab}}/U_{\mathrm{diam}}=\frac{\delta}{2}\sum_\ell q_\ell^{-1}$
does not depend on the weights or the features, so it can be computed from
the graphs alone. We evaluate it for depths $L\in\{2,4,8\}$ on
Cora~\cite{sen2008collective} and MUTAG~\cite{debnath1991structure}, whose
maximum degrees differ by two orders of magnitude. MUTAG is a low-degree
setting in which GIN performs well on graph
classification~\cite{xu2018powerful}; we apply the bound to its node
representations before pooling, on the disjoint union of 20 test graphs. We
consider four edge-deletion strategies, each removing 10\% of the edges:
random deletion, deletion within high-$k$ cores, deletion of
high-betweenness edges, and adversarial deletion with
PR-BCD~\cite{geisler2021robustness}; for betweenness, we also delete 50\%.

\begin{table}[t]
\centering
\small
\begin{tabular}{lccccc}
\toprule
Perturbation & $\delta$ & $r$ ($L=2$) & $r$ ($L=4$) & $r$ ($L=8$) & $L_{\mathrm{crit}}$ \\
\midrule
\multicolumn{6}{l}{\textit{Cora, node classification} ($d_{\max}=168$, $q=169$)} \\
random 10\,\%      & 4.23  & 0.025 & 0.050 & 0.100 & 80 \\
PR-BCD 10\,\%      & 3.77  & 0.022 & 0.045 & 0.089 & 90 \\
$k$-core 10\,\%    & 9.20  & 0.054 & 0.109 & 0.218 & 37 \\
betweenness 10\,\% & 7.50  & 0.044 & 0.089 & 0.178 & 46 \\
betweenness 50\,\% & 12.92 & 0.076 & 0.153 & 0.306 & 27 \\
\midrule
\multicolumn{6}{l}{\textit{MUTAG, graph classification} ($d_{\max}=3$, $q=4$)} \\
random 10\,\%      & 1.27 & 0.32 & 0.64 & \textbf{1.27} & 7 \\
PR-BCD 10\,\%      & 1.41 & 0.35 & 0.71 & \textbf{1.41} & 6 \\
$k$-core 10\,\%    & 1.55 & 0.39 & 0.78 & \textbf{1.55} & 6 \\
betweenness 10\,\% & 1.66 & 0.41 & 0.83 & \textbf{1.66} & 5 \\
betweenness 50\,\% & 2.44 & 0.61 & \textbf{1.22} & \textbf{2.44} & 4 \\
\bottomrule
\end{tabular}
\caption{Ratio $r$ of the GIN stability bound (Theorem~\ref{thm:gin}) to the
diameter bound (Lemma~\appref{lem:gin_diameter}{8}), with $\epsilon_\ell=0$
and $\delta$ averaged over three seeds. The bound is informative iff $r<1$;
$r\ge1$ in bold. $L_{\mathrm{crit}}=\lceil 2q/\delta\rceil$ is the first
depth at which informativeness is lost. For MUTAG, $\delta$ is the largest
per-graph perturbation norm, which equals the operator norm on the
block-diagonal union.}
\label{tab:regime_dmax}
\end{table}

\paragraph{Results.}
Table~\ref{tab:regime_dmax} shows the two regimes predicted by
Corollary~\ref{cor:gin_informativeness}. On Cora, whose maximum degree is
168, the bound is informative by a wide margin for every strategy and depth
tested, with $r\le0.31$, and remains so up to depths of 27 to 90 layers. On
MUTAG, whose maximum degree is 3, the margin is small: the bound stops being
informative at depth 8 for every strategy, and already at depth 4 when half
of the high-betweenness edges are removed. The strategies also differ in the
size of the perturbation they induce for the same number of deleted edges.
Deletions that target the core of the graph ($k$-core, betweenness) produce
the largest $\delta$, whereas PR-BCD, although adversarial for the
downstream task, produces the smallest one on Cora. Informativeness is thus
governed by where edges are removed, not only by how many, and a
perturbation designed to degrade predictions need not be large in operator
norm. Finally, informativeness
is relative: both bounds are worst-case, and $r<1$ means that the stability
bound adds information over the diameter bound, not that it is tight.

\section{Conclusion and Future Work}
We derived structural stability bounds for GINs and single-head GATv1 and
compared them with perturbation-independent diameter bounds. For GINs, the
stability bound is informative exactly when the perturbation is small
relative to the ratio of maximum degree to depth, a condition met with a
wide margin on Cora but violated beyond a few layers on MUTAG. For
GATv1, the bound never is, because the structural term charges every changed
attention row its largest possible change. Two directions follow: sharper
GATv1 bounds based on the attention mass actually assigned to added and
removed neighbors, of which \extref\ gives a first step, and
extensions to multi-head attention. More broadly, we suggest that stability
bounds be reported together with the trivial bound they must beat.

\begin{footnotesize}

\bibliographystyle{unsrt}
\bibliography{biblio}

\end{footnotesize}

% ****************************************************************************
% END OF BIBLIOGRAPHY AREA
% ****************************************************************************

% ****************************************************************************
% APPENDIX (extended arXiv version only)
% ****************************************************************************
\newpage
\appendix

% Appendix of the extended version (proofs). Input by main_extended.tex.
% \input{proof_gcn}

\section{Stability of GIN}

\paragraph{Graph and perturbation.}
Let $\mS,\widehat{\mS}\in\{0,1\}^{N\times N}$ be the adjacency matrices of two undirected, simple graphs defined on the same $N$ vertices. We assume
\[
\mS=\mS^\top,\qquad \widehat{\mS}=\widehat{\mS}^\top,
\qquad \mS_{ii}=\widehat{\mS}_{ii}=0.
\]
Self-information is incorporated explicitly in the GIN aggregation, so self-loops are \emph{not} included in $\mS$ or $\widehat{\mS}$. We use $\fr{\cdot}$ for the Frobenius norm and $\op{\cdot}$ for the Euclidean operator (spectral) norm. Define
\begin{equation}\label{eq:degree}
d_{\max}
:=\max\!\left\{
\max_i\sum_j \mS_{ij},\;
\max_i\sum_j\widehat{\mS}_{ij}
\right\}.
\end{equation}
The features $\mX\in\mathbb{R}^{N\times d_0}$ and all neural-network parameters are identical for the two graphs. In particular, structural perturbations do not modify $\mX$ or the learned parameters.

\paragraph{An explicit MLP at every GIN layer.}
For $\ell=0,\ldots,L-1$, choose an integer $K_\ell\geq 1$, widths
$ p_{\ell,0}=d_\ell,\ldots,p_{\ell,K_\ell}=d_{\ell+1}$,
weight matrices and bias vectors
\[
\mW_{\ell,j}\in\mathbb{R}^{p_{\ell,j-1}\times p_{\ell,j}},
\qquad
\vb_{\ell,j}\in\mathbb{R}^{p_{\ell,j}},
\quad j=1,\ldots,K_\ell.
\]
Let $\rho_{\ell,j}$ be an elementwise $1$-Lipschitz activation with
$\rho_{\ell,j}(0)=0$; the identity mapping is permitted, for instance at the output layer. In particular, ReLU satisfies these assumptions. For an input matrix $\mZ\in\mathbb{R}^{N\times d_\ell}$, define
\begin{align}
\mZ_{\ell,0}&=\mZ,\nonumber\\
\mZ_{\ell,j}
&=\rho_{\ell,j}\!\left(
\mZ_{\ell,j-1}\mW_{\ell,j}
+\vone_N \vb_{\ell,j}^\top
\right),\quad j=1,\ldots,K_\ell,\label{eq:mlp_def}\\
\mathrm{MLP}_\ell(\mZ)&=\mZ_{\ell,K_\ell}.\nonumber
\end{align}
Here $\vone_N\in\mathbb{R}^N$ is the all-ones vector. Thus the same MLP is applied independently to each node's aggregated features; it does not mix node indices internally.

Define two explicit constants determined solely by MLP parameters:
\begin{align}
\lambda_\ell
&:=\prod_{j=1}^{K_\ell}\op{\mW_{\ell,j}},
\label{eq:lambda}\\
\beta_\ell
&:=\sum_{j=1}^{K_\ell}
\lVert \vb_{\ell,j}\rVert_2
\prod_{r=j+1}^{K_\ell}\op{\mW_{\ell,r}}.
\label{eq:beta}
\end{align}
All empty products in this document are defined to equal $1$.

\paragraph{The $L$-layer GIN.}
For a fixed scalar $\varepsilon_\ell\in\mathbb{R}$ at each layer, define
\begin{equation}\label{eq:propagator}
\mP_\ell(\mT):=(1+\varepsilon_\ell)\mI_N+\mT,
\qquad \mT\in\{\mS,\widehat{\mS}\}.
\end{equation}
The GIN is the recursion
\begin{equation}\label{eq:gin_def}
\mH^{(0)}(\mT)=\mX,\qquad
\mH^{(\ell+1)}(\mT)
=\mathrm{MLP}_\ell\!\left(\mP_\ell(\mT)\mH^{(\ell)}(\mT)\right),
\quad \ell=0,\ldots,L-1,
\end{equation}
with output $\Phi(\mT,\mX):=\mH^{(L)}(\mT)$. This is precisely the sum-aggregating GIN architecture, with an explicitly defined MLP. No batch-normalization, input-dependent rescaling, or stochastic dropout is assumed.

For later use, set
\begin{equation}\label{eq:q_delta}
q_\ell:=|1+\varepsilon_\ell|+d_{\max},\qquad
\delta:=\op{\mS-\widehat{\mS}},\qquad
\delta_F:=\fr{\mS-\widehat{\mS}}.
\end{equation}
In particular, $\delta\leq\delta_F$.

\subsection{Preliminary lemmas}

\begin{lemma}[Lipschitz constant and growth of the MLP]
\label{lem:mlp}
For all $\mU,\mV\in\mathbb{R}^{N\times d_\ell}$,
\begin{equation}\label{eq:mlp_lipschitz}
\fr{\mathrm{MLP}_\ell(\mU)-\mathrm{MLP}_\ell(\mV)}
\leq \lambda_\ell\fr{\mU-\mV}.
\end{equation}
Moreover, for all $\mU\in\mathbb{R}^{N\times d_\ell}$,
\begin{equation}\label{eq:mlp_growth}
\fr{\mathrm{MLP}_\ell(\mU)}
\leq \lambda_\ell\fr{\mU}+\sqrt{N}\,\beta_\ell.
\end{equation}
\end{lemma}

\begin{proof}
For matrices of the same shape, entrywise $1$-Lipschitz continuity implies
\[
\fr{\rho_{\ell,j}(\mC)-\rho_{\ell,j}(\mD)}\leq\fr{\mC-\mD}.
\]
Consequently, for the intermediate representations corresponding to $\mU$ and $\mV$,
\[
\fr{\mZ_{\ell,j}(\mU)-\mZ_{\ell,j}(\mV)}
\leq
\op{\mW_{\ell,j}}
\fr{\mZ_{\ell,j-1}(\mU)-\mZ_{\ell,j-1}(\mV)},
\]
since the two bias matrices cancel. Iterating for $j=1,\ldots,K_\ell$ proves \eqref{eq:mlp_lipschitz}.

To bound the image of zero, let $f_\ell$ denote the same MLP applied to a single row vector. Starting from the zero vector, $\rho_{\ell,j}(0)=0$ and the triangle inequality give
\[
\lVert f_{\ell,j}(0)\rVert_2
\leq
\op{\mW_{\ell,j}}\lVert f_{\ell,j-1}(0)\rVert_2
+\lVert \vb_{\ell,j}\rVert_2.
\]
Unrolling the recursion yields
$\lVert f_\ell(0)\rVert_2\leq\beta_\ell$.
Because $\mathrm{MLP}_\ell(0)$ has $N$ identical rows,
$\fr{\mathrm{MLP}_\ell(0)}=\sqrt N\lVert f_\ell(0)\rVert_2
\leq\sqrt N\beta_\ell$.
Finally,
\[
\begin{aligned}
\fr{\mathrm{MLP}_\ell(\mU)}
&\leq\fr{\mathrm{MLP}_\ell(\mU)-\mathrm{MLP}_\ell(0)}
     +\fr{\mathrm{MLP}_\ell(0)}\\
&\leq\lambda_\ell\fr{\mU}+\sqrt N\beta_\ell,
\end{aligned}
\]
which is \eqref{eq:mlp_growth}.
\end{proof}

\begin{lemma}[Uniform bound on the aggregation operator]
\label{lem:prop}
For $\mT\in\{\mS,\widehat{\mS}\}$ and every layer $\ell$,
\begin{equation}\label{eq:prop_bound}
\op{\mP_\ell(\mT)}\leq q_\ell.
\end{equation}
Furthermore,
\begin{equation}\label{eq:prop_diff}
\mP_\ell(\mS)-\mP_\ell(\widehat{\mS})=\mS-\widehat{\mS}.
\end{equation}
\end{lemma}

\begin{proof}
As $\mT$ is a symmetric, nonnegative adjacency matrix of maximum degree at most $d_{\max}$,
\[
\op{\mT}^2\leq\lVert \mT\rVert_1\lVert \mT\rVert_\infty
\leq d_{\max}^2.
\]
The triangle inequality therefore implies
\[
\op{\mP_\ell(\mT)}
\leq |1+\varepsilon_\ell|+\op{\mT}
\leq |1+\varepsilon_\ell|+d_{\max}=q_\ell.
\]
Equation \eqref{eq:prop_diff} follows by cancellation of the common diagonal term.
\end{proof}

\begin{lemma}[Explicit bounds on intermediate node representations]
\label{lem:representation}
Let $B_0:=\fr{\mX}$ and, for $\ell=0,\ldots,L-1$, define
\begin{equation}\label{eq:B_recurrence}
B_{\ell+1}:=\lambda_\ell q_\ell B_\ell+\sqrt N\beta_\ell.
\end{equation}
Then, for both $\mT=\mS$ and $\mT=\widehat{\mS}$,
$\fr{\mH^{(\ell)}(\mT)}\leq B_\ell$. Moreover, the bound is explicit:
\begin{equation}\label{eq:B_explicit}
\begin{aligned}
B_\ell={}&
\fr{\mX}\prod_{r=0}^{\ell-1}(\lambda_rq_r)\\
&+\sqrt N\sum_{j=0}^{\ell-1}\beta_j
\prod_{r=j+1}^{\ell-1}(\lambda_rq_r).
\end{aligned}
\end{equation}
\end{lemma}

\begin{proof}
Using Lemmas~\ref{lem:mlp} and \ref{lem:prop},
\[
\begin{aligned}
\fr{\mH^{(\ell+1)}(\mT)}
&\leq\lambda_\ell\fr{\mP_\ell(\mT)\mH^{(\ell)}(\mT)}
     +\sqrt N\beta_\ell\\
&\leq\lambda_\ell q_\ell\fr{\mH^{(\ell)}(\mT)}
     +\sqrt N\beta_\ell.
\end{aligned}
\]
Since $\mH^{(0)}(\mT)=\mX$, induction gives
$\fr{\mH^{(\ell)}(\mT)}\leq B_\ell$.
To verify \eqref{eq:B_explicit}, note that it holds for $\ell=0$, where all sums vanish and all products are empty. Substituting it into \eqref{eq:B_recurrence} yields the same expression with $\ell$ replaced by $\ell+1$, thereby completing the induction.
\end{proof}

\subsection{Complete structural stability guarantee}

\begin{theorem}[Explicit stability bound for a biased $L$-layer GIN]
\label{thm:main}
Consider the GIN \eqref{eq:gin_def} under the assumptions above. Then its output representations satisfy
\begin{equation}\label{eq:main_simple}
\begin{aligned}
&\fr{\Phi(\mS,\mX)-\Phi(\widehat{\mS},\mX)}\\
&\quad\leq
\op{\mS-\widehat{\mS}}
\sum_{t=0}^{L-1}\lambda_t B_t
\prod_{r=t+1}^{L-1}(\lambda_rq_r)\\
&\quad\leq
\fr{\mS-\widehat{\mS}}
\sum_{t=0}^{L-1}\lambda_t B_t
\prod_{r=t+1}^{L-1}(\lambda_rq_r),
\end{aligned}
\end{equation}
where $B_t$ has the fully explicit expression \eqref{eq:B_explicit}. In particular, substituting the latter gives a bound with no unknown intermediate representations:
\begin{equation}\label{eq:main_explicit}
\begin{aligned}
&\fr{\Phi(\mS,\mX)-\Phi(\widehat{\mS},\mX)}
\leq \fr{\mS-\widehat{\mS}}\\
&\quad\times\sum_{t=0}^{L-1}\lambda_t
\left[
\fr{\mX}\prod_{r=0}^{t-1}(\lambda_rq_r)
+\sqrt N\sum_{j=0}^{t-1}\beta_j
   \prod_{r=j+1}^{t-1}(\lambda_rq_r)
\right]\\
&\qquad\qquad\times
\prod_{r=t+1}^{L-1}(\lambda_rq_r).
\end{aligned}
\end{equation}
All constants depend only on $L$, $\mX$, the MLP weights and biases, the GIN self-weights $\varepsilon_\ell$, and $d_{\max}$; the right-hand side vanishes whenever $\mS=\widehat{\mS}$.
\end{theorem}

\begin{proof}
For $\ell=0,\ldots,L$, write
\[
\mH_\ell:=\mH^{(\ell)}(\mS),\qquad
\widehat \mH_\ell:=\mH^{(\ell)}(\widehat{\mS}),\qquad
\Delta_\ell:=\fr{\mH_\ell-\widehat \mH_\ell}.
\]
The common input features imply $\Delta_0=0$. By Lemma~\ref{lem:mlp} and the definition of the GIN,
\begin{equation}\label{eq:step1}
\begin{aligned}
\Delta_{\ell+1}
&=\fr{\mathrm{MLP}_\ell(\mP_\ell(\mS)\mH_\ell)
 -\mathrm{MLP}_\ell(\mP_\ell(\widehat{\mS})\widehat \mH_\ell)}\\
&\leq\lambda_\ell
\fr{\mP_\ell(\mS)\mH_\ell
 -\mP_\ell(\widehat{\mS})\widehat \mH_\ell}.
\end{aligned}
\end{equation}
Add and subtract $\mP_\ell(\widehat{\mS})\mH_\ell$ inside the norm. The resulting identity is
\begin{equation}\label{eq:decomp}
\begin{aligned}
&\mP_\ell(\mS)\mH_\ell
-\mP_\ell(\widehat{\mS})\widehat \mH_\ell\\
&\quad=\bigl(\mP_\ell(\mS)-\mP_\ell(\widehat{\mS})\bigr)\mH_\ell
+\mP_\ell(\widehat{\mS})(\mH_\ell-\widehat \mH_\ell)\\
&\quad=(\mS-\widehat{\mS})\mH_\ell
+\mP_\ell(\widehat{\mS})(\mH_\ell-\widehat \mH_\ell).
\end{aligned}
\end{equation}
Using the triangle inequality, submultiplicativity
$\fr{AB}\leq\op{A}\fr{B}$,
and Lemmas~\ref{lem:prop} and \ref{lem:representation}, we infer
\begin{equation}\label{eq:drift_recursion}
\begin{aligned}
\Delta_{\ell+1}
&\leq\lambda_\ell\left(
\op{\mS-\widehat{\mS}}\fr{\mH_\ell}
+\op{\mP_\ell(\widehat{\mS})}\Delta_\ell
\right)\\
&\leq\lambda_\ell\delta B_\ell
+\lambda_\ell q_\ell\Delta_\ell.
\end{aligned}
\end{equation}
This recurrence is \emph{closed}: $B_\ell$ is explicitly specified by \eqref{eq:B_explicit}, and no perturbed representation appears on the right-hand side.

We next solve the recurrence. For $m\in\{1,\ldots,L\}$, repeated substitution gives
\begin{equation}\label{eq:drift_unroll}
\Delta_m
\leq
\delta\sum_{t=0}^{m-1}\lambda_t B_t
\prod_{r=t+1}^{m-1}(\lambda_rq_r).
\end{equation}
Indeed, for $m=1$ this follows from \eqref{eq:drift_recursion} and $\Delta_0=0$. If it holds for $m$, then
\[
\begin{aligned}
\Delta_{m+1}
&\leq \lambda_m q_m\Delta_m+\lambda_m\delta B_m\\
&\leq\delta\sum_{t=0}^{m-1}\lambda_t B_t
\prod_{r=t+1}^{m}(\lambda_rq_r)
+\delta\lambda_mB_m,
\end{aligned}
\]
which is the desired expression for $m+1$ (the last product is empty). Taking $m=L$ proves the first inequality in \eqref{eq:main_simple}. The second follows from
$\op{\mS-\widehat{\mS}}\leq\fr{\mS-\widehat{\mS}}$.
Finally, inserting \eqref{eq:B_explicit} yields \eqref{eq:main_explicit}, completing the proof.
\end{proof}

\subsection{Simplifications and interpretation}

\begin{corollary}[Bias-free MLPs: compact closed-form stability bound]
\label{cor:biasfree}
If $\vb_{\ell,j}=0$ for all $\ell,j$ (and hence $\beta_\ell=0$), then
\begin{equation}\label{eq:biasfree_exact}
\begin{aligned}
&\fr{\Phi(\mS,\mX)-\Phi(\widehat{\mS},\mX)}\\
&\quad\leq
\op{\mS-\widehat{\mS}}\fr{\mX}
\left(\prod_{\ell=0}^{L-1}\lambda_\ell\right)
\sum_{t=0}^{L-1}
\prod_{\substack{r=0\\r\ne t}}^{L-1}q_r.
\end{aligned}
\end{equation}
In particular, with $q:=\max_\ell q_\ell$,
\begin{equation}\label{eq:final_simple}
\begin{aligned}
&\fr{\Phi(\mS,\mX)-\Phi(\widehat{\mS},\mX)}\\
&\quad\leq
Lq^{L-1}\fr{\mX}
\left(\prod_{\ell=0}^{L-1}\lambda_\ell\right)
\op{\mS-\widehat{\mS}}.
\end{aligned}
\end{equation}
Since $\op{\mS-\widehat{\mS}}\le\fr{\mS-\widehat{\mS}}$, both bounds also
hold with the Frobenius norm.
\end{corollary}

\begin{proof}
When all biases vanish, the explicit representation bound is
$B_t=\fr{\mX}\prod_{r=0}^{t-1}(\lambda_rq_r)$.
The $t$-th summand in \eqref{eq:main_simple} therefore equals
\[
\fr{\mX}\left(\prod_{r=0}^{L-1}\lambda_r\right)
\prod_{\substack{r=0\\r\ne t}}^{L-1}q_r.
\]
Summing over $t$ proves \eqref{eq:biasfree_exact}.
Each product in that sum contains exactly $L-1$ factors bounded by $q$, which gives \eqref{eq:final_simple}.
\end{proof}

In particular, Theorem~\ref{thm:gin} in the main text is
\eqref{eq:biasfree_exact}.

\begin{corollary}[Explicit bound under $B$ undirected edge flips]
\label{cor:budget}
Suppose $\mS$ and $\widehat{\mS}$ differ by exactly $B$ undirected edge additions or deletions, with each edge counted once. Then
\[
\fr{\mS-\widehat{\mS}}=\sqrt{2B}.
\]
Consequently, the conclusion of Theorem~\ref{thm:main} holds with
$\fr{\mS-\widehat{\mS}}$ replaced by $\sqrt{2B}$.
In particular, in the bias-free setting,
\begin{equation}\label{eq:budget}
\begin{aligned}
&\fr{\Phi(\mS,\mX)-\Phi(\widehat{\mS},\mX)}\\
&\quad\leq
\sqrt{2B}\,Lq^{L-1}\fr{\mX}
\prod_{\ell=0}^{L-1}\lambda_\ell.
\end{aligned}
\end{equation}
\end{corollary}

\begin{proof}
Each undirected edge flip changes exactly two symmetric adjacency entries by $\pm1$. Hence the squared Frobenius norm of the adjacency difference is $2B$. Apply Theorem~\ref{thm:main} or the Frobenius-norm form of Corollary~\ref{cor:biasfree}.
\end{proof}

\paragraph{Remark (A bound using only the original maximum degree).}
If $d_{\max}(\mS):=\max_i\sum_j \mS_{ij}$ and at most $B$ edges are flipped, then $d_{\max}\leq d_{\max}(\mS)+B$. Thus one may choose
\[
q_\ell=|1+\varepsilon_\ell|+d_{\max}(\mS)+B
\]
in all upper bounds. This provides a completely pre-attack bound in terms of $\mS$, the budget $B$, and the trained network.

\paragraph{Remark (Scope of the theorem).}
The bound is global and does not assume differentiability of ReLU or any linearization of the GIN. It is a worst-case Lipschitz-type guarantee for the \emph{node representation matrix}; additional readout or pooling layers require their own Lipschitz factors. It explicitly accounts for ordinary affine MLP biases. A GIN employing batch normalization, dropout, edge attributes, or graph-dependent parameters requires additional assumptions and cannot be covered automatically by the constants above. The factor $q^{L-1}$ can grow rapidly with $L$ and the maximum degree; the result is a valid upper bound, not necessarily a tight one.

\section{Representation diameter of GINs}

\begin{lemma}[Representation diameter of GINs]
\label{lem:gin_diameter}
Under the assumptions of Theorem~\ref{thm:gin},
let $\mathcal{S}$ denote the set of adjacency matrices
of undirected graphs without self-loops on the same
labelled vertex set, with maximum degree at most
$d_{\max}$.

For fixed input features $\mX$ and fixed model
parameters, define the representation set
\[
\mathcal{R}_{\Phi}(\mX)
=
\left\{
\Phi(\mS,\mX):\mS\in\mathcal{S}
\right\}.
\]

Then its diameter, measured in Frobenius norm,
satisfies
\[
\operatorname{diam}(\mathcal{R}_{\Phi}(\mX))
\leq
2\left(\prod_{\ell=0}^{L-1}
\lambda_\ell q_\ell\right)\fr{\mX}.
\]
\end{lemma}

\begin{proof}
For any $\mS\in\mathcal{S}$, the GIN update is
\[
\mX^{(\ell+1)}
=
\operatorname{MLP}_{\ell}
\left(
(\mS+(1+\epsilon_\ell)\mI)
\mX^{(\ell)}
\right).
\]

Since each MLP is $\lambda_\ell$-Lipschitz
and vanishes at zero,
\[
\fr{\mX^{(\ell+1)}}
\leq
\lambda_\ell
\op{\mS+(1+\epsilon_\ell)\mI}
\fr{\mX^{(\ell)}}.
\]

Since $\mS$ is symmetric and its maximum degree
is at most $d_{\max}$,
\[
\op{\mS}\leq d_{\max}.
\]

Consequently,
\[
\op{\mS+(1+\epsilon_\ell)\mI}
\leq |1+\epsilon_\ell|+d_{\max}
=q_\ell.
\]

Iterating over all layers yields
\[
\fr{\Phi(\mS,\mX)}
\leq
\left(\prod_{\ell=0}^{L-1}
\lambda_\ell q_\ell\right)\fr{\mX}.
\]

For any $\mS,\hat{\mS}\in\mathcal{S}$,
the triangle inequality gives
\[
\begin{aligned}
&\fr{\Phi(\mS,\mX)-\Phi(\hat{\mS},\mX)}\\
&\quad\leq
\fr{\Phi(\mS,\mX)}
+\fr{\Phi(\hat{\mS},\mX)}\\
&\quad\leq
2\left(\prod_{\ell=0}^{L-1}
\lambda_\ell q_\ell\right)\fr{\mX}.
\end{aligned}
\]

Taking the supremum over all pairs
$\mS,\hat{\mS}\in\mathcal{S}$
concludes the proof.
\end{proof}

\begin{proof}[Proof of Corollary~\ref{cor:gin_informativeness}]
By Theorem~\ref{thm:gin}, the stability
upper bound is
\[
B_{\mathrm{stab}}
=
\delta
\left(\prod_{\ell=0}^{L-1}\lambda_\ell\right)
\left(
\sum_{t=0}^{L-1}
\prod_{\substack{r=0\\r\neq t}}^{L-1}q_r
\right)\fr{\mX}.
\]

Since all $q_\ell$ are positive,
\[
B_{\mathrm{stab}}
=
\delta
\left(\prod_{\ell=0}^{L-1}
\lambda_\ell q_\ell\right)
\left(\sum_{t=0}^{L-1}\frac{1}{q_t}\right)
\fr{\mX}.
\]

By Lemma~\ref{lem:gin_diameter}, the diameter
upper bound is
\[
B_{\mathrm{diam}}
=
2\left(\prod_{\ell=0}^{L-1}
\lambda_\ell q_\ell\right)\fr{\mX}.
\]

Dividing by the common positive factor yields
\[
B_{\mathrm{stab}}<B_{\mathrm{diam}}
\iff
\delta\sum_{\ell=0}^{L-1}\frac{1}{q_\ell}<2.
\]

When all $q_\ell=q$, this becomes
$\delta L/q<2$, concluding the proof.
\end{proof}

\section{Stability analysis of Graph Attention Networks}

\paragraph{Setting and notation.}
Let $L\geq 1$, and let
$S,\widehat S\in\{0,1\}^{N\times N}$
be the adjacency matrices of two undirected graphs
without self-loops on the same labelled vertex set:
\[
S=S^\top,\qquad
\widehat S=\widehat S^\top,\qquad
S_{ii}=\widehat S_{ii}=0
\quad\text{for all }i.
\]
For $T\in\{S,\widehat S\}$, define
\[
\mathcal N_T(i)
:=
\{j\in\{1,\ldots,N\}:T_{ij}=1\}\cup\{i\}.
\]
Thus, self-loops are included in the attention
neighborhoods, rather than in the adjacency matrices.
This neighborhood coincides with
$\mathcal N_T^+(i)$ in Definition~\ref{def:gat}.

Set
\[
d_i(T):=|\mathcal N_T(i)|,
\qquad
d_{\max}
:=
\max_{T\in\{S,\widehat S\}}
\max_{1\leq i\leq N}d_i(T).
\]
In particular, every attention neighborhood is nonempty.

We consider an $L$-layer, single-head, bias-free
GATv1 network.
The two evaluations use the same input
$X\in\mathbb R^{N\times F^{(0)}}$
and the same trainable parameters
\[
W^{(\ell)}
\in\mathbb R^{F^{(\ell)}\times F^{(\ell+1)}},
\qquad
a^{(\ell)}\in\mathbb R^{2F^{(\ell+1)}},
\qquad \ell=0,\ldots,L-1.
\]
No normalization layers or stochastic dropout
are included.

Throughout this appendix, $\|\cdot\|_{\mathrm{op}}$
denotes the Euclidean operator norm,
$\|\cdot\|_F$ the Frobenius norm, and
$\|\cdot\|_2$ the Euclidean vector norm.
For a matrix $H$ with rows $H_i$, write
\[
\|H\|_{2,\infty}:=\max_i\|H_i\|_2.
\]
Empty products equal one and empty sums equal zero.

\paragraph{Attention and propagation.}
For $H\in\mathbb R^{N\times F^{(\ell)}}$, define
the attention logits by
\[
g_{ij}^{(\ell)}(H)
:=
\operatorname{LeakyReLU}\!\left(
(a^{(\ell)})^\top
\left[
(H_iW^{(\ell)})^\top
\,\Vert\,
(H_jW^{(\ell)})^\top
\right]
\right),
\]
where $H_i$ is the $i$th row of $H$ and
$\Vert$ denotes concatenation.
The negative slope of LeakyReLU belongs to $[0,1]$.

For $T\in\{S,\widehat S\}$, define the attention matrix
$P^{(\ell)}(T,H)\in\mathbb R^{N\times N}$ by
\[
P_{ij}^{(\ell)}(T,H)
:=
\begin{cases}
\displaystyle
\frac{\exp(g_{ij}^{(\ell)}(H))}
{\sum_{k\in\mathcal N_T(i)}
 \exp(g_{ik}^{(\ell)}(H))},
& j\in\mathcal N_T(i),\\[2ex]
0, & \text{otherwise}.
\end{cases}
\]
Each attention matrix is nonnegative and row-stochastic:
\[
P_{ij}^{(\ell)}(T,H)\geq 0,
\qquad
\sum_{j=1}^N P_{ij}^{(\ell)}(T,H)=1.
\]

The node representations satisfy
\[
H^{(0)}(T)=X,
\qquad
H^{(\ell+1)}(T)
=
\sigma\!\left(
P^{(\ell)}(T,H^{(\ell)}(T))
H^{(\ell)}(T)W^{(\ell)}
\right),
\]
and the network output is
\[
\Phi(T,X):=H^{(L)}(T).
\]
The activation $\sigma$ is applied elementwise,
is $1$-Lipschitz, and satisfies $\sigma(0)=0$.
The arguments below also allow different activations
at different layers, provided that each satisfies
these assumptions.

\paragraph{Elementary norm bounds.}

\begin{lemma}[Lipschitz continuity of activations]
For matrices $U,V$ of the same dimensions,
\[
\|\sigma(U)-\sigma(V)\|_F
\leq \|U-V\|_F.
\]
Moreover,
\[
\|\sigma(U)\|_F\leq \|U\|_F.
\]
The same inequalities hold in Euclidean norm
for individual rows.
LeakyReLU with negative slope in $[0,1]$
is also $1$-Lipschitz.
\end{lemma}

\begin{proof}
The scalar Lipschitz inequality applies entrywise.
Summing its square over the entries gives the
Frobenius-norm bound.
The growth bound follows by taking $V=0$
and using $\sigma(0)=0$.
The row-wise statement follows identically.
Finally, LeakyReLU is continuous and piecewise linear,
with slopes in $[0,1]$, so it is $1$-Lipschitz.
\end{proof}

\begin{lemma}[Norm of the attention matrix]
\label{lemma:attention_matrix_norm}
For every layer $\ell$, every
$T\in\{S,\widehat S\}$, and every
$H\in\mathbb R^{N\times F^{(\ell)}}$,
\[
\|P^{(\ell)}(T,H)\|_{\mathrm{op}}
\leq \sqrt{d_{\max}},
\qquad
\|P^{(\ell)}(T,H)\|_F
\leq \sqrt N.
\]
\end{lemma}

\begin{proof}
Write $P=P^{(\ell)}(T,H)$.
For the induced matrix norms,
\[
\|P\|_\infty:=\max_i\sum_j|P_{ij}|,
\qquad
\|P\|_1:=\max_j\sum_i|P_{ij}|,
\]
we have
\[
\|P\|_{\mathrm{op}}^2
\leq \|P\|_1\|P\|_\infty.
\]
Since $P$ is nonnegative and row-stochastic,
$\|P\|_\infty=1$.
Since $T$ is undirected,
\[
j\in\mathcal N_T(i)
\quad\Longleftrightarrow\quad
i\in\mathcal N_T(j).
\]
Thus, column $j$ has at most $d_j(T)$ nonzero entries.
Each entry is at most one, so
\[
\|P\|_1
=
\max_j\sum_{i:\,j\in\mathcal N_T(i)}P_{ij}
\leq d_{\max}.
\]
This proves the operator-norm bound.

For the Frobenius norm, $0\leq P_{ij}\leq 1$ gives
\[
\|P\|_F^2
=
\sum_{i,j}P_{ij}^2
\leq
\sum_{i,j}P_{ij}
=N.
\]
\end{proof}

\begin{lemma}[Frobenius norm of node representations]
\label{lemma:gat_embedding_bound}
For every $T\in\{S,\widehat S\}$ and
$\ell\in\{0,\ldots,L\}$,
\[
\|H^{(\ell)}(T)\|_F
\leq
d_{\max}^{\ell/2}\|X\|_F
\prod_{t=0}^{\ell-1}
\|W^{(t)}\|_{\mathrm{op}}.
\]
At the first layer, we also have
\[
\|H^{(1)}(T)\|_F
\leq
\sqrt N\,\|X\|_{\mathrm{op}}
\|W^{(0)}\|_{\mathrm{op}}.
\]
\end{lemma}

\begin{proof}
The activation bound and
Lemma~\ref{lemma:attention_matrix_norm} imply
\begin{align*}
\|H^{(\ell+1)}(T)\|_F
&\leq
\|P^{(\ell)}(T,H^{(\ell)}(T))
H^{(\ell)}(T)W^{(\ell)}\|_F\\
&\leq
\sqrt{d_{\max}}\,
\|H^{(\ell)}(T)\|_F
\|W^{(\ell)}\|_{\mathrm{op}}.
\end{align*}
Iteration from $H^{(0)}(T)=X$ proves the first bound.
For the alternative first-layer estimate,
\begin{align*}
\|H^{(1)}(T)\|_F
&\leq
\|P^{(0)}(T,X)\|_F
\|X\|_{\mathrm{op}}
\|W^{(0)}\|_{\mathrm{op}}\\
&\leq
\sqrt N\,\|X\|_{\mathrm{op}}
\|W^{(0)}\|_{\mathrm{op}}.
\end{align*}
\end{proof}

\paragraph{One-step propagation of perturbations.}
For brevity, write
\[
H_\ell:=H^{(\ell)}(S),
\qquad
\widehat H_\ell:=H^{(\ell)}(\widehat S),
\]
and
\[
P_\ell:=P^{(\ell)}(S,H_\ell),
\qquad
\widehat P_\ell
:=P^{(\ell)}(\widehat S,\widehat H_\ell).
\]
Define
\[
\Delta_\ell:=\|\widehat H_\ell-H_\ell\|_F,
\qquad
\Gamma_\ell:=\|\widehat P_\ell-P_\ell\|_F,
\qquad
w_\ell:=\|W^{(\ell)}\|_{\mathrm{op}}.
\]
The common input implies $\Delta_0=0$.

\begin{theorem}[One-step perturbation bound]
\label{thm:gat_recursive_stability}
For every $\ell=0,\ldots,L-1$,
\[
\Delta_{\ell+1}
\leq
w_\ell\left(
\sqrt{d_{\max}}\,\Delta_\ell
+
\|H_\ell\|_{\mathrm{op}}\Gamma_\ell
\right).
\]
\end{theorem}

\begin{proof}
By the Lipschitz continuity of $\sigma$,
\[
\Delta_{\ell+1}
\leq
\|\widehat P_\ell\widehat H_\ell W^{(\ell)}
-P_\ell H_\ell W^{(\ell)}\|_F.
\]
Using
\[
\widehat P_\ell\widehat H_\ell-P_\ell H_\ell
=
\widehat P_\ell(\widehat H_\ell-H_\ell)
+
(\widehat P_\ell-P_\ell)H_\ell,
\]
the triangle inequality gives
\begin{align*}
\Delta_{\ell+1}
&\leq
w_\ell\left(
\|\widehat P_\ell\|_{\mathrm{op}}\Delta_\ell
+
\Gamma_\ell\|H_\ell\|_{\mathrm{op}}
\right)\\
&\leq
w_\ell\left(
\sqrt{d_{\max}}\,\Delta_\ell
+
\|H_\ell\|_{\mathrm{op}}\Gamma_\ell
\right),
\end{align*}
where the last step uses
Lemma~\ref{lemma:attention_matrix_norm}.
\end{proof}

\paragraph{Separating feature and structural effects.}
Adding and subtracting
$P^{(\ell)}(\widehat S,H_\ell)$ yields
\[
\Gamma_\ell\leq F_\ell+R_\ell,
\]
where
\begin{align*}
F_\ell
&:=
\|P^{(\ell)}(\widehat S,\widehat H_\ell)
-P^{(\ell)}(\widehat S,H_\ell)\|_F,\\
R_\ell
&:=
\|P^{(\ell)}(\widehat S,H_\ell)
-P^{(\ell)}(S,H_\ell)\|_F.
\end{align*}
The term $F_\ell$ measures attention variation
due to changes in representations at fixed topology.
The term $R_\ell$ measures attention variation
due to changes in topology at fixed representations.

\begin{lemma}[Euclidean Lipschitz constant of softmax]
\label{lem:gat_softmax}
For every integer $m\geq 1$ and every
$z,z'\in\mathbb R^m$,
\[
\|\operatorname{softmax}(z)
-\operatorname{softmax}(z')\|_2
\leq \frac12\|z-z'\|_2.
\]
\end{lemma}

\begin{proof}
Let $p=\operatorname{softmax}(z)$.
Its Jacobian is
\[
J(z)=\operatorname{diag}(p)-pp^\top.
\]
This matrix is symmetric and positive semidefinite.
The diagonal entry in row $i$ is $p_i(1-p_i)$,
and the sum of the absolute values of the
off-diagonal entries in that row is also
$p_i(1-p_i)$.
Gershgorin's theorem therefore gives
\[
\|J(z)\|_{\mathrm{op}}
=
\lambda_{\max}(J(z))
\leq 2\max_i p_i(1-p_i)
\leq \frac12.
\]
The integral identity
\[
\operatorname{softmax}(z')
-\operatorname{softmax}(z)
=
\int_0^1
J(z+t(z'-z))(z'-z)\,dt
\]
proves the result.
\end{proof}

\begin{lemma}[Attention variation at a fixed graph]
\label{lem:gat_fixed_graph_attention}
For every $T\in\{S,\widehat S\}$ and
$H,H'\in\mathbb R^{N\times F^{(\ell)}}$,
\begin{equation}
\label{eq:gat_feature_attention_bound}
\|P^{(\ell)}(T,H')-P^{(\ell)}(T,H)\|_F
\leq c_\ell\|H'-H\|_F,
\end{equation}
where
\[
c_\ell
:=
\sqrt{\frac{d_{\max}}2}\,
\|a^{(\ell)}\|_2w_\ell.
\]
\end{lemma}

\begin{proof}
Fix a layer and suppress its superscript.
Write $D=H'-H$ and partition the attention vector as
\[
a=
\begin{bmatrix}
a_L\\a_R
\end{bmatrix}.
\]
For every $j\in\mathcal N_T(i)$,
the Lipschitz continuity of LeakyReLU and
Cauchy--Schwarz imply
\[
|g_{ij}(H')-g_{ij}(H)|
\leq
\|W\|_{\mathrm{op}}
\left(
\|a_L\|_2\|D_i\|_2
+
\|a_R\|_2\|D_j\|_2
\right).
\]
Using $(x+y)^2\leq 2x^2+2y^2$ gives
\begin{align*}
&\sum_i\sum_{j\in\mathcal N_T(i)}
|g_{ij}(H')-g_{ij}(H)|^2\\
&\quad\leq
2\|W\|_{\mathrm{op}}^2
\left(
\|a_L\|_2^2\sum_i d_i(T)\|D_i\|_2^2
+
\|a_R\|_2^2\sum_j d_j(T)\|D_j\|_2^2
\right)\\
&\quad\leq
2d_{\max}\|W\|_{\mathrm{op}}^2
\|a\|_2^2\|D\|_F^2.
\end{align*}
Here, the second degree count uses the symmetry
of the attention-neighborhood relation.

For each row $i$, both attention vectors are softmax
outputs on the same index set $\mathcal N_T(i)$.
By Lemma~\ref{lem:gat_softmax},
their squared Euclidean distance is at most one
quarter of the squared distance between their logits.
Padding the attention vectors with zeros outside
$\mathcal N_T(i)$ does not change this distance.
Summing over rows therefore yields
\[
\|P^{(\ell)}(T,H')-P^{(\ell)}(T,H)\|_F^2
\leq
\frac{d_{\max}}2
w_\ell^2\|a^{(\ell)}\|_2^2
\|H'-H\|_F^2.
\]
Taking square roots proves the claim.
\end{proof}

\begin{lemma}[Attention variation under binary adjacency changes]
\label{lem:gat_structure_attention}
For every fixed representation matrix
$H\in\mathbb R^{N\times F^{(\ell)}}$,
\begin{equation}
\label{eq:gat_structure_attention_bound}
\|P^{(\ell)}(\widehat S,H)-P^{(\ell)}(S,H)\|_F
\leq
\sqrt2\,\|\widehat S-S\|_F.
\end{equation}
\end{lemma}

\begin{proof}
For two probability vectors $p,q\in\mathbb R^N$,
nonnegativity gives $p^\top q\geq 0$, and hence
\[
\|p-q\|_2^2
=
\|p\|_2^2+\|q\|_2^2-2p^\top q
\leq 2.
\]
If $S_{i,:}=\widehat S_{i,:}$, then the attention
neighborhoods in row $i$ coincide.
Since $H$ and the trainable parameters are fixed,
the corresponding attention rows are identical.
Otherwise, their squared distance is at most two.
Consequently,
\begin{align*}
&\|P^{(\ell)}(\widehat S,H)
-P^{(\ell)}(S,H)\|_F^2\\
&\quad\leq
2\sum_{i=1}^N
\mathbf 1_{\{\widehat S_{i,:}\neq S_{i,:}\}}\\
&\quad\leq
2\sum_{i,j}(\widehat S_{ij}-S_{ij})^2.
\end{align*}
The last inequality holds because the adjacency
matrices are binary: each changed row contains
at least one entry whose squared difference is one.
Taking square roots proves the bound.
\end{proof}

\paragraph{Uniform control of representation magnitudes.}
Define
\[
B_\ell
:=
\|X\|_{2,\infty}
\prod_{t=0}^{\ell-1}w_t,
\qquad \ell=0,\ldots,L.
\]

\begin{lemma}[Degree-independent row-norm bound]
\label{lem:gat_row_bound}
For every $T\in\{S,\widehat S\}$ and
$\ell=0,\ldots,L$,
\[
\|H^{(\ell)}(T)\|_{2,\infty}\leq B_\ell,
\qquad
\|H^{(\ell)}(T)\|_{\mathrm{op}}
\leq \sqrt N\,B_\ell.
\]
\end{lemma}

\begin{proof}
Since $\sigma$ is nonexpansive and vanishes at zero,
and each attention row defines a convex combination,
\begin{align*}
\|H_i^{(\ell+1)}(T)\|_2
&\leq
\left\|
\sum_j
P_{ij}^{(\ell)}(T,H^{(\ell)}(T))
H_j^{(\ell)}(T)W^{(\ell)}
\right\|_2\\
&\leq
\sum_j
P_{ij}^{(\ell)}(T,H^{(\ell)}(T))
\|H_j^{(\ell)}(T)W^{(\ell)}\|_2\\
&\leq
w_\ell\|H^{(\ell)}(T)\|_{2,\infty}.
\end{align*}
Taking the maximum over $i$ and iterating from
$H^{(0)}(T)=X$ proves the first assertion.
The second follows from
\[
\|H\|_{\mathrm{op}}
\leq \|H\|_F
\leq \sqrt N\,\|H\|_{2,\infty}.
\]
\end{proof}

\paragraph{Global structural stability.}

\begin{theorem}[Global structural stability of single-head GATv1]
\label{thm:gat_complete_stability}
Under the assumptions above, define, for
$\ell=0,\ldots,L-1$,
\begin{align}
q_\ell
&:=
w_\ell\left(
\sqrt{d_{\max}}
+
\sqrt{\frac{Nd_{\max}}2}\,
B_\ell\|a^{(\ell)}\|_2w_\ell
\right),
\label{eq:gat_q}\\
\gamma_\ell
&:=
\sqrt{2N}\,w_\ell B_\ell.
\label{eq:gat_b}
\end{align}
Then
\begin{equation}
\label{eq:gat_global_stability}
\begin{aligned}
\|\Phi(\widehat S,X)-\Phi(S,X)\|_F
&\leq
C_{\mathrm{GAT}}\|\widehat S-S\|_F,\\
C_{\mathrm{GAT}}
&:=
\sum_{t=0}^{L-1}
\gamma_t\prod_{r=t+1}^{L-1}q_r.
\end{aligned}
\end{equation}
\end{theorem}

\begin{proof}
Set $\delta:=\|\widehat S-S\|_F$.
The decomposition $\Gamma_\ell\leq F_\ell+R_\ell$,
together with
Lemmas~\ref{lem:gat_fixed_graph_attention}
and~\ref{lem:gat_structure_attention}, gives
\[
\Gamma_\ell
\leq c_\ell\Delta_\ell+\sqrt2\,\delta.
\]
By Theorem~\ref{thm:gat_recursive_stability}
and Lemma~\ref{lem:gat_row_bound},
\begin{align*}
\Delta_{\ell+1}
&\leq
w_\ell\left(
\sqrt{d_{\max}}\,\Delta_\ell
+
\sqrt N\,B_\ell
(c_\ell\Delta_\ell+\sqrt2\,\delta)
\right)\\
&=
q_\ell\Delta_\ell+\gamma_\ell\delta.
\end{align*}
We now solve this recurrence.
Since $\Delta_0=0$, induction gives, for
$m=1,\ldots,L$,
\[
\Delta_m
\leq
\delta
\sum_{t=0}^{m-1}
\gamma_t\prod_{r=t+1}^{m-1}q_r.
\]
Indeed, the case $m=1$ follows directly from
$\Delta_1\leq\gamma_0\delta$.
If the bound holds for $m<L$, then
\begin{align*}
\Delta_{m+1}
&\leq q_m\Delta_m+\gamma_m\delta\\
&\leq
\delta\sum_{t=0}^{m-1}
\gamma_t\prod_{r=t+1}^{m}q_r
+\gamma_m\delta\\
&=
\delta\sum_{t=0}^{m}
\gamma_t\prod_{r=t+1}^{m}q_r.
\end{align*}
Taking $m=L$ proves
\eqref{eq:gat_global_stability}.
\end{proof}

\paragraph{Refinement using attention masses.}
Fix a layer $\ell$ and a representation matrix
$H\in\mathbb R^{N\times F^{(\ell)}}$.
For each node $i$, define
\[
C_i:=\mathcal N_S(i)\cap\mathcal N_{\widehat S}(i),
\]
\[
D_i^-:=\mathcal N_S(i)\setminus\mathcal N_{\widehat S}(i),
\qquad
D_i^+:=\mathcal N_{\widehat S}(i)\setminus\mathcal N_S(i).
\]
The attention masses assigned to removed and
added neighbors are, respectively,
\[
r_i:=
\sum_{j\in D_i^-}P_{ij}^{(\ell)}(S,H),
\qquad
u_i:=
\sum_{j\in D_i^+}P_{ij}^{(\ell)}(\widehat S,H).
\]

\begin{lemma}[Exact row-wise total-variation identity]
\label{lem:gat_mass_refinement}
Let
\[
p_i:=P_{i,:}^{(\ell)}(S,H),
\qquad
\widehat p_i:=P_{i,:}^{(\ell)}(\widehat S,H).
\]
With the convention
$\operatorname{TV}(p,q):=\tfrac12\|p-q\|_1$,
we have
\[
\operatorname{TV}(p_i,\widehat p_i)
=
\max\{r_i,u_i\}.
\]
Equivalently,
\[
\|p_i-\widehat p_i\|_1
=
2\max\{r_i,u_i\}.
\]
Furthermore,
\[
\|P^{(\ell)}(\widehat S,H)
-P^{(\ell)}(S,H)\|_F
\leq
\left(
2\sum_{i=1}^N\max\{r_i,u_i\}^2
\right)^{1/2}.
\]
\end{lemma}

\begin{proof}
Fix $i$.
On $C_i$, the unnormalized attention weights
are identical for the two graphs, because
$H$ and the trainable parameters are fixed.
The restrictions of $p_i$ and $\widehat p_i$
to $C_i$ are therefore proportional.
Their differences have a common sign on $C_i$,
so
\[
\sum_{j\in C_i}
|(p_i)_j-(\widehat p_i)_j|
=
|(1-r_i)-(1-u_i)|
=
|r_i-u_i|.
\]
On $D_i^-$, only $p_i$ can be nonzero,
with total mass $r_i$.
On $D_i^+$, only $\widehat p_i$ can be nonzero,
with total mass $u_i$.
Both vanish outside the union of the neighborhoods.
Thus,
\[
\|p_i-\widehat p_i\|_1
=
r_i+u_i+|r_i-u_i|
=
2\max\{r_i,u_i\}.
\]

Since both vectors have total mass one,
the positive and negative parts of
$p_i-\widehat p_i$ each have $\ell_1$ norm
$\max\{r_i,u_i\}$.
These parts have disjoint supports.
Applying $\|v\|_2\leq\|v\|_1$ to each part yields
\[
\|p_i-\widehat p_i\|_2^2
\leq 2\max\{r_i,u_i\}^2.
\]
Summing over rows proves the matrix bound.
\end{proof}

\paragraph{Scope of the result.}
The global theorem concerns changes in binary
adjacency matrices on a fixed labelled vertex set,
with identical input features and trainable parameters.
Self-loops are included in every attention neighborhood.
The structural estimate is a bound on a discrete
domain, rather than a continuity statement for
arbitrary real-valued adjacency masks.

The constants may depend on the input magnitude,
the network parameters, $N$, $d_{\max}$, and depth.
The result does not require an additional assumed
uniform bound on attention logits.
Lemma~\ref{lem:gat_mass_refinement} can replace
the uniform structural estimate when the attention
masses of changed edges are available.

The theorem controls node representations before
graph-level readout or classification.
Biases, multiple attention heads, and normalization
layers require additional analysis.

\begin{lemma}[Representation diameter of GATs]
\label{lem:gat_diameter}
Under the assumptions of Theorem~\ref{thm:gatv1},
let $\mathcal{S}$ denote the set of adjacency
matrices of undirected graphs without self-loops
on the same labelled vertex set.

For fixed input features $\mX$ and fixed model
parameters, define the representation set
\[
\mathcal{R}_{\Phi}(\mX)
=
\left\{
\Phi(\mS,\mX):\mS\in\mathcal{S}
\right\}.
\]

Then its diameter, measured in Frobenius norm,
satisfies
\[
\operatorname{diam}(\mathcal{R}_{\Phi}(\mX))
\leq 2\sqrt{N}\,B_L,
\]
where
\[
B_L
=
\|\mX\|_{2,\infty}
\prod_{\ell=0}^{L-1}w_\ell.
\]
\end{lemma}

\begin{proof}
Let $\mH^{(0)}=\mX$ and denote by
$\mH^{(\ell)}$ the node representations
at layer $\ell$.

For any $\mS\in\mathcal{S}$, the GATv1 update
at node $i$ is
\[
\mH_i^{(\ell+1)}
=
\sigma\left(
\sum_{j\in\mathcal{N}_S^+(i)}
\alpha_{ij}^{(\ell)}
\mW^{(\ell)}\mH_j^{(\ell)}
\right).
\]

The attention coefficients satisfy
\[
\alpha_{ij}^{(\ell)}\geq 0,
\qquad
\sum_{j\in\mathcal{N}_S^+(i)}
\alpha_{ij}^{(\ell)}=1.
\]

Since $\sigma$ is $1$-Lipschitz and
$\sigma(0)=0$,
\[
\begin{aligned}
\|\mH_i^{(\ell+1)}\|_2
&\leq
\left\|
\sum_{j\in\mathcal{N}_S^+(i)}
\alpha_{ij}^{(\ell)}
\mW^{(\ell)}\mH_j^{(\ell)}
\right\|_2\\
&\leq
w_\ell
\sum_{j\in\mathcal{N}_S^+(i)}
\alpha_{ij}^{(\ell)}
\|\mH_j^{(\ell)}\|_2\\
&\leq
w_\ell\|\mH^{(\ell)}\|_{2,\infty}.
\end{aligned}
\]

Taking the maximum over all nodes gives
\[
\|\mH^{(\ell+1)}\|_{2,\infty}
\leq
w_\ell\|\mH^{(\ell)}\|_{2,\infty}.
\]

By induction,
\[
\|\mH^{(\ell)}\|_{2,\infty}
\leq
\|\mX\|_{2,\infty}
\prod_{t=0}^{\ell-1}w_t
=B_\ell.
\]

In particular,
\[
\|\Phi(\mS,\mX)\|_F
\leq
\sqrt{N}\,
\|\Phi(\mS,\mX)\|_{2,\infty}
\leq
\sqrt{N}\,B_L.
\]

For any $\mS,\hat{\mS}\in\mathcal{S}$,
the triangle inequality yields
\[
\begin{aligned}
&\fr{\Phi(\mS,\mX)-\Phi(\hat{\mS},\mX)}\\
&\quad\leq
\fr{\Phi(\mS,\mX)}
+
\fr{\Phi(\hat{\mS},\mX)}\\
&\quad\leq 2\sqrt{N}\,B_L.
\end{aligned}
\]

Taking the supremum over all pairs
$\mS,\hat{\mS}\in\mathcal{S}$
concludes the proof.
\end{proof}

\begin{proof}[Proof of Lemma~\ref{lem:gat_informativeness}]
By Theorem~\ref{thm:gatv1}, the stability
upper bound is
\[
B_{\mathrm{stab}}
=
\delta
\sum_{t=0}^{L-1}
\gamma_t
\prod_{r=t+1}^{L-1}q_r.
\]

Using
\[
\gamma_t
=
\sqrt{2N}\,w_tB_t
=
\sqrt{2N}\,B_{t+1}
\]
and
\[
B_{t+1}
=
B_L
\left(
\prod_{r=t+1}^{L-1}w_r
\right)^{-1},
\]
we obtain
\[
B_{\mathrm{stab}}
=
\delta\sqrt{2N}\,B_L
\sum_{t=0}^{L-1}
\prod_{r=t+1}^{L-1}
\frac{q_r}{w_r}.
\]

By Lemma~\ref{lem:gat_diameter}, the
diameter upper bound is
\[
B_{\mathrm{diam}}
=
2\sqrt{N}\,B_L.
\]

Since $B_L>0$, comparison gives
\[
\frac{B_{\mathrm{stab}}}
{B_{\mathrm{diam}}}
=
\frac{\delta}{\sqrt{2}}
\sum_{t=0}^{L-1}
\prod_{r=t+1}^{L-1}
\frac{q_r}{w_r}.
\]

Therefore,
\[
B_{\mathrm{stab}}<B_{\mathrm{diam}}
\iff
\delta
\sum_{t=0}^{L-1}
\prod_{r=t+1}^{L-1}
\frac{q_r}{w_r}
<\sqrt{2}.
\]

We now show that this condition cannot
hold for distinct admissible graphs.

Since $\mS$ and $\hat{\mS}$ are symmetric,
binary adjacency matrices with zero diagonal,
\[
\begin{aligned}
\delta^2
&=
\sum_{i,j=1}^{N}
(S_{ij}-\hat S_{ij})^2\\
&=
2\sum_{1\leq i<j\leq N}
(S_{ij}-\hat S_{ij})^2.
\end{aligned}
\]

For distinct graphs, at least one
undirected edge differs. Hence,
\[
\delta^2\geq 2,
\qquad
\delta\geq\sqrt{2}.
\]

Moreover, all $q_r$ and $w_r$ are positive,
and the term corresponding to $t=L-1$
is an empty product equal to one.
Thus,
\[
\sum_{t=0}^{L-1}
\prod_{r=t+1}^{L-1}
\frac{q_r}{w_r}
\geq 1.
\]

Consequently,
\[
\frac{B_{\mathrm{stab}}}
{B_{\mathrm{diam}}}
\geq
\frac{\delta}{\sqrt{2}}
\geq 1.
\]

Therefore, the stability upper bound
cannot be strictly smaller than the
diameter upper bound for any nonzero
admissible perturbation.
\end{proof}

\end{document}